# Metasurface-enhanced Light Detection and Ranging Technology

**Authors**

Renato Juliano Martins[1], Emil Marinov[1], M. Aziz Ben Youssef[1], Christina Kyrou[1], Mathilde Joubert[1], Constance Colmagro[1,2], Valentin Gâté[2], Colette Turbil[2], Pierre-Marie Coulon[1], Daniel Turover[2], Samira Khadir[1], Massimo Giudici[3], Charalambos Klitis[4], Marc Sorel[4,5] and Patrice Genevet[1] †

**Affiliations**

[1] Université Cote d'Azur, CNRS, CRHEA, Rue Bernard Gregory, Sophia Antipolis 06560 Valbonne, France

[2] NAPA-Technologies, 74160 Archamps, France

[3] Université Côte d'Azur, Centre National de La Recherche Scientifique, Institut de Physique de Nice, F-06560 Valbonne, France

[4] School of Engineering, University of Glasgow, Glasgow, G12 8LT, UK

[5] Institute of Technologies for Communication, Information and Perception (TeCIP), Sant'Anna School of Advanced Studies, Via Moruzzi 1, 56127, Pisa, Italy

† Corresponding Author: Patrice.Genevet@crhea.cnrs.fr

**Abstract:**

**Deploying advanced imaging solutions to robotic and autonomous systems by mimicking human vision requires simultaneous acquisition of multiple fields of views, named the peripheral and fovea regions. Low-resolution peripheral field provides coarse scene exploration to direct the eye to focus to a highly resolved fovea region for sharp imaging. Among 3D computer vision techniques, Light Detection and Ranging (LiDAR) is currently considered at the industrial level for robotic vision. LiDAR is an imaging technique that monitors pulses of light at optical frequencies to sense the space and to recover three-dimensional ranging information. Notwithstanding the efforts on LiDAR integration and optimization, commercially available devices have slow frame rate and low image resolution, notably limited by the performance of mechanical or slow solid-state deflection systems. Metasurfaces (MS) are versatile optical components that can distribute the optical power in desired regions of space. Here, we report on an advanced LiDAR technology that uses ultrafast low FoV deflectors cascaded with large area metasurfaces to achieve large FoV and simultaneous peripheral and central imaging zones. This technology achieves MHz frame rate for 2D imaging, and up to KHz for 3D imaging, with extremely large FoV (up to 150°deg. on both vertical and horizontal scanning axes). The use of this disruptive LiDAR technology with advanced learning algorithms offers perspectives to improve further the perception capabilities and decision-making process of autonomous vehicles and robotic systems.**

**Introduction**

Autonomous mobile systems such as autonomous cars and warehouse robots include multiple sensors to acquire information of their surrounding environments, defining their position, velocity, and acceleration in real time. Among them, range sensors, and in particular the optical ranging sensors, provide vision to robotic systems[1–3] and are thus at the core of the automation of industrial processes, the so-called 4.0 industrial revolution. Several optical imaging techniques are currently integrated into industrial robots for 3D image acquisition, including stereoscopic camera, RADAR, structured light illumination and laser range finders or LiDARs. LiDAR is a technological concept introduced in the early 60s, when Massachusetts Institute of Technology (MIT) scientists reported on the detection of echo signals upon sending optical radiation to the moon surface[4]. Since the pioneering MIT work, LiDARs have been using laser sources to illuminate targeted objects and to collect the returning echo signals offering the possibility of reconstructing highly resolved three-dimensional (3D) images. Conventional LiDARs rely on Time-of-Flight (ToF) measurement, which employs a pulsed laser directed towards a distant reflective object to measure the round-trip time of light pulses propagating from the laser to the scanned scene and back to a detection module. All LiDAR components must act synchronously to tag single returning pulses for ranging imaging reconstruction. The formula, $2d = c\,T_{oF}$, holds for the recovered distance, where $c$ is the speed of light and $T_{oF}$ is the ToF. To sense the space, the LiDAR source must be able to sweep a large Field of View (FoV). The objects in the scene are then detected, point-by-point by measuring the ToF from every single direction to build an optical echo map. The other measurement processes known as Amplitude Modulation Continuous Wave (AMCW)[5,6], Frequency Modulation Continuous Wave (FMCW)[7,8] or Stepped Frequency Continuous Wave (SFCW)[9] employ continuous waves with constant or time-modulated frequency to measure the round-trip time of the modulated light information. LiDAR systems enable the real-time 3D mapping of objects located at long, medium or short-range distance from the source, finding a vast variety of applications beyond robotic vision, spanning from landscape mapping Chase[10–12], atmospheric particle detection[13–16], wind speed measurement[17,18], static and/or moving object tracking[19–22], AR/VR[23], among others. Generally, LiDARs are classified into scanning or non-scanning (Flash LiDAR) systems depending on whether the laser sources simply illuminate[26] or scan the targeted scene. A scanning LiDAR system can be essentially described in terms of three key components, (i) the light source for illumination, (ii) the scanning module for fast beam direction at different points in the scene, and (iii) the detection system for high-speed recovering of the optical information received from the scene. Over the past decades, nanophotonics-based LiDAR systems have blossomed, and new advanced scanning and detection techniques have been proposed[24,25]. The expected massive use of LiDARs in the automotive industry for advanced driver assistance systems (ADAS) or even full-autonomous driving brought out new challenges for the scanning systems, including low fabrication complexity, potential for scalable manufacturing, cost, lightweight, tolerance to vibrations and so on. Today, industrially relevant LiDARs mainly use macro-mechanical systems to scan the entire 360º FoV. Besides their large FoV, these bulk systems present limited imaging rates of the order of few tens of Hz. A promising evolution in mechanical scanners are the micro-electromechanical systems[27] (MEMS) which shift the scanning frequency to the kHz range. However, a major drawback of MEMS is the low FoV typically not exceeding 25º for horizontal and 15º for vertical scanning. At the research level, beam steering with optical phased arrays (OPA)[28,29] provides remarkable speeds while reaching FoV around 60º. However, OPA technology is less likely to be massively deployed in industrial systems due to its manufacturing challenges. The industrially mature liquid crystal modulators are also not adequate as LiDAR scanners due to their poor FoV usually remaining below 20º depending on the wavelength, as well as their kHz modulation frequency[30,31]. Moreover, acousto-optic deflectors (AODs) enabling ultrafast MHz scanning[32,33], have never been considered in LiDARs because of their narrow FoV reaching at maximum 2º, imposing a compromise between high-speed imaging and large FoV.

During the last decade, metasurfaces (MS)[34] have spurred the interest of the entire international photonic community by unveiling the possibility of engineering the properties (i.e., the amplitude, the phase, the frequency and/or the polarization) of light at will[35]. They are flat optical components made of arrangements of scattering objects (meta-atoms) of subwavelength size and periodicity. Currently, four light modulation mechanisms are used to create metasurfaces: light scattering from resonant nanoparticles[36,37], geometric phase occurring during polarization conversion (Pancharatnam-Berry phase)[38], accumulated propagation phase in pillars with controllable effective Refractive Index (ERI)[39] and the topological phase in vicinity of singularities[40]. Usually, MSs comprise inherently passive components, designed to perform a fixed optical functionality after fabrication. For instance, by properly selecting the size and the spacing of the meta-atoms, MSs allow to redirect a laser beam at any arbitrary but fixed angle dictated by the generalized Snell's law. Clearly, passive MS alone cannot be used in LiDARs requiring real-time beam scanning. On the contrary, dynamic MSs designed by - or combined with - materials possessing tunable optical properties caused by external stimuli[41–45] stand as promising alternatives for real-time deflection. Recently, the US startup company LUMOTIVE introduced electrically addressable reflective resonant MSs infiltrated with liquid crystals and demonstrated scanning LiDARs of MHz frequency and FoV of around $120°$[46]. The latter approach has been proven extremely auspicious for miniaturized, scalable LiDARs but it involves complex electronic architectures, and likely significant optical losses associated with the metallic MS building blocks.

Here, we propose an alternative high-frequency beam scanning approach that exploits the light deflecting capabilities of passive MSs to expand the LiDAR FoV to 150x150°, and to achieve simultaneous low- and high-resolution multi-zones imaging. We make use of an ERI multibeam deflecting MS cascaded with a commercial AOD. The system offers large flexibilities in terms of beam scanning performance, operation wavelength and materials. The angular resolution, referring to the ability of the system to distinguish adjacent targets and retrieve shapes, becomes very important in applications requiring simultaneous long and short-range detections. Our multi-zone LiDAR imaging demonstration can mimic human vision by achieving simultaneous high frame rate acquisition of high and low field zones with different spatial resolution. The large design flexibility of MSs provides imaging capabilities of interest to LiDAR systems, meanwhile offering new industrial applications.

## Ultra-fast and high FoV metasurface scanning module

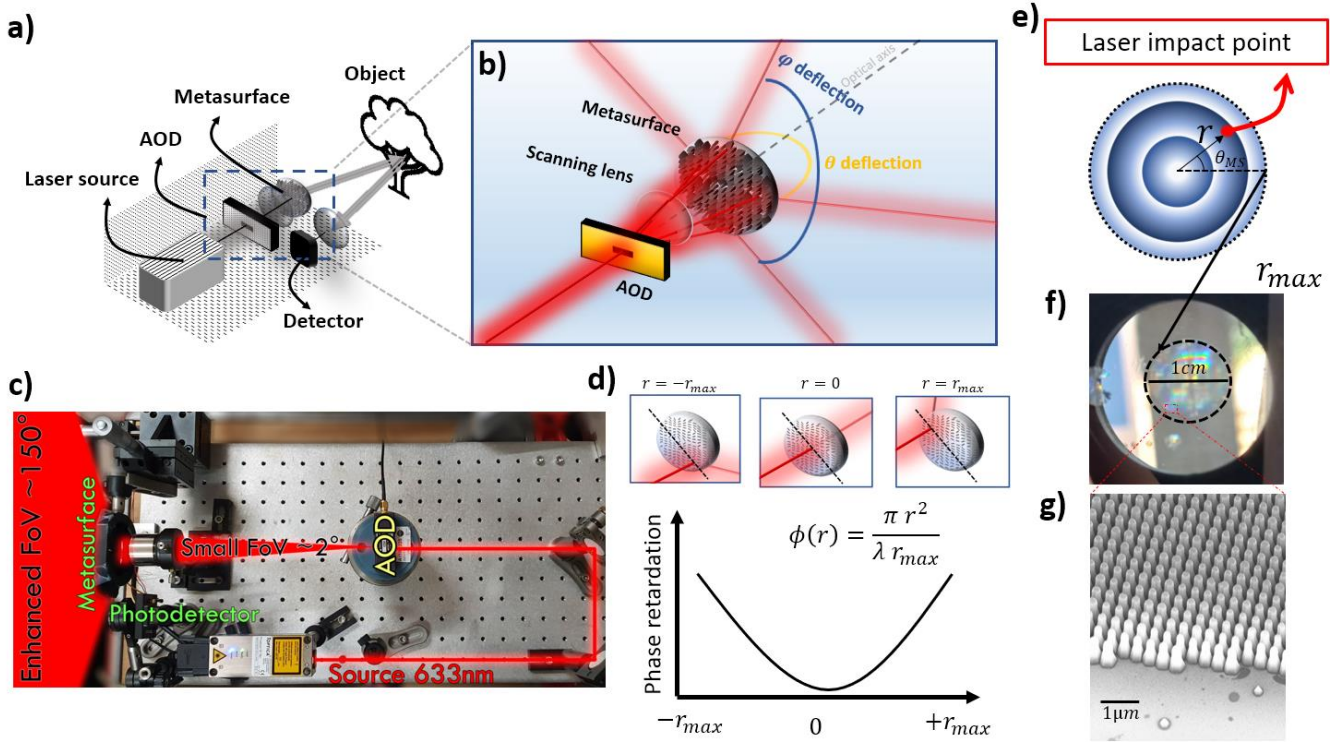

*Figure 1 – **a)** Schematic representation of the LIDAR system. A triggered laser source, emitting single pulses for ToF detection, is directed to a synchronized acousto-optic deflector offering ultrafast light scanning with low FoV (~2°). The deflected beam is directed to a scanning lens to scan the laser spot on the metasurface at different radial and azimuthal positions. The transmitted light across the metasurface is deviated according to the position of the impinging beam on the component to cover a scanning range between −75° and 75°. The scattered light from the scene is collected using a fast detector. Data are processed to extract the single echo ToF for 2D and 3D imaging of the scene. **b)** Detail of the cascaded AOD-metasurface assemblied deflection system. **c)** Top view photography of the optical setup. **d)** (bottom) Graphical representation of the metasurface phase distribution along the radial axis. (top) Representations of beam deflection according to the incident beam positioning on the metasurface. **e)** IllustrationI of axial symmetry for the laser impact point. **f)** Photography of the 1cm MS fabricated using nanoimprinting lithography. **g)** SEM image of the sample showing the nanopillar buiding blocks of varying sizes employed to achieve beam deflection by considering lateral effective refractive index variations.*

MHz beam scanning can be achieved over extreme FoV, by coupling AODs with ERI MSs exhibiting spatially varying deflection angles. Figure 1a) illustrates the experimental concept where a modulated laser source at $\lambda = 633\ nm$ (TOPTICA i-beam smart) generates single pulses at any arbitrary rate up to $250\ MHz$. For single pulse LIDAR, the repetition rate $f_{rep}$ is related to the maximum ranging distance $d_{max}$ by the expression:

$$d_{max} = \frac{1}{2}\frac{c}{f_{rep}}.$$

The focused beam with a small deflection is angularly increased to scan in both azimuthal $\theta$ and elevation $\varphi$ angles. A detailed scheme of the FoV amplifying system is shown in Figure 1b). A photography of the built proof-of-concept system is shown in Figure 1c) where we highlighted (shaded red region) the expansion of the small two degrees (2°) AOD FoV into an enhanced 150° FoV. The deflected angle by the MS is controlled by the impact position of the impinging focused beam on the MS plane, associated with the radial and angular coordinates $r$ and $\theta_{MS}$, respectively (see Figure 1e)). By applying voltage into the AOD, one can actively re-point the beam at any arbitrary angle within the 2°x2°FoV, thus sweeping the focused beam across the metasurface to vary $\theta_{MS}$ and $r$, in the range of $[0 − 2\pi]$ and $[0 − r_{max}]$, respectively, where $r_{max}$ is the radius of the metasurface. Note that

$\theta_{MS}$ and $r$ denote, in polar coordinate, the position of the impact beam on the metasurface according to fig. 1e). For simplicity in connecting incident and deflected angles, we designed a circular metasurface with radially symmetric phase delaying response, but given the versatility in controlling the optical wavefront, various MS with any other beam defecting properties can be adjusted according to specific application. We must also highlight that, in principle, there is no limitation on the observed FoV as it is fully dependent on the metasurface phase function, within the limit $[0, \pi]$ for transmission scheme. In this first demonstration, we implemented (Figures 1f)-g)) the simple concept of ERI MS designed to spatially impart linearly increasing momentum with respect to the radial dimension $r$ given by the expression:

$$\frac{\partial \Phi}{\partial r} = -k_0 \frac{r}{r_{max}} \quad [1]$$

Where, $k_0$ is the free space momentum, and $\Phi$ the local phase retardation. Such design results in parabolic phase retardation as represented in Figure 1d). In this design, the deflected beam will be delayed by a maximum phase retardation of $\Phi = \mp \frac{\pi r_{max}}{\lambda}$ and $\Phi = 0$ for the peripherical points $\pm r_{max}$ and central points, respectively. Moreover, equation [1] transformed in Cartesian coordinates determines the value of the deflected angles in both axes, denoted as $(\theta, \varphi)$, according to the generalized Snell laws[41]:

$$\begin{cases} k_{x,t} = k_{x,i} + \frac{\partial \Phi}{\partial x} = k_0 \sin \theta_i \sin \varphi_i + \frac{\partial \Phi}{\partial r} \frac{\partial r}{\partial x} \\ k_{y,t} = k_{y,i} + \frac{\partial \Phi}{\partial y} = k_0 \sin \theta_i \cos \varphi_i + \frac{\partial \Phi}{\partial r} \frac{\partial r}{\partial y} \end{cases} , \quad [2]$$

where the phase gradient is defined at the metasurface plane at $z = 0$. Considering small incident angles originating from the AOD, the expressions simplify as:

$$\begin{cases} k_0 \sin \theta_t \sin \varphi_t = -k_0 \frac{r}{r_{max}} \cos \theta_{MS} \\ k_0 \sin \theta_t \cos \varphi_t = -k_0 \frac{r}{r_{max}} \sin \theta_{MS} \end{cases} , \quad [3]$$

Such expression validates the linearity observed for small angles [-40°,40°] according to the experimental measurements of the voltage dependence of the deflection angles supplemental materials Figure S2c).

**2D and 3D LiDAR image acquisition**

To show the angular and depth 2D imaging capabilities of our LIDAR system, we first performed 1D scanning of three distinct objects placed on a table, (1) a square reflector mounted on a post, (2) a round deflector and (3) a box reflector, angularly distributed at different locations as shown in Figure 2a). The associated 2D LIDAR ranging image is displayed in Figure 2b), indicating that high reflectivity objects are observed at LIDAR positions matching to those observed with a conventional camera (Figure 2a). Particularly, we found that the three objects shown in Figure 2c) were located at the following width [x], and depth [z] positions: [-0.4m, 1.5m], [-0.1m, 2.4m] and [0.6m, 3.5m] for the square, the round, and the box reflector respectively. In the graph, we also observe the difference in reflectivity of the three objects at various distances leading to distinct intensities: the objects on the left and right (square and box deflectors) correspond to lower signals due to their angular locations, size and distance, while the round deflector in the middle has higher reflectivity and appears with higher reflectance. This first example validates the short-range (~5 m) imaging capabilities of our LiDAR system.

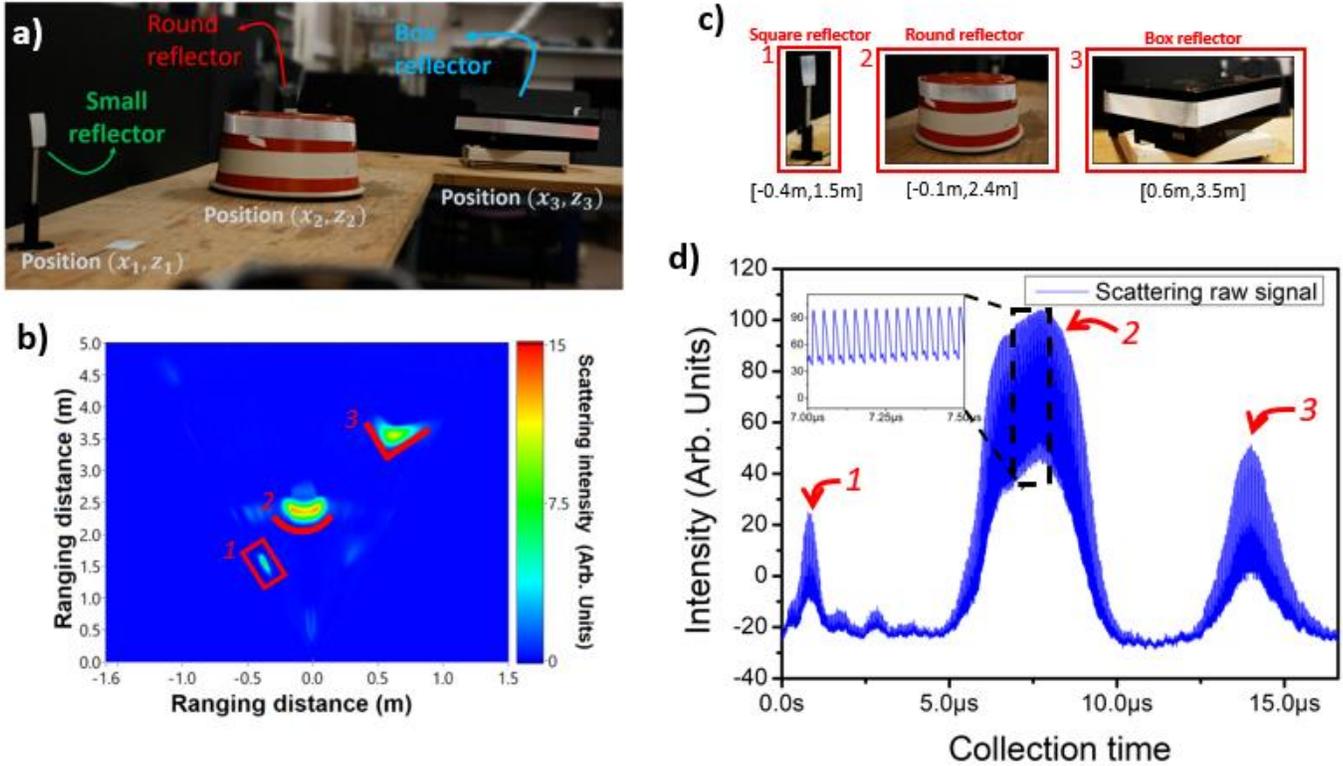

*Figure 2 – **a)** Photography of the scene **b)** Ranging image of three objects displaced on a table using high reflectives tapes to improve the intensity of the returned signal. In 1 a post with a small reflector was used in 2 a round object with a reflector and in 3 there is a box reflector with a tape around it. The graph shows the image in the correct ranging distance X (scanning dimension) and Z (ranging dimension) showing the capabilities to sense all of the three objects. **c)** Position of single objects according to ranging image in b). And **d)** Raw signal collected for the respective image, showing that objects oriented in the normal direction have bigger scattering intensity, the inset display single pulses used to determine the ToF ranging distance.*

To further investigate the capabilities of the system, we extended the performance to achieve 3D imaging. To this end, an additional FoV dimension is added by cascading a second AOD, orthogonally oriented, in the elevation axis. The extended FoV is now improved over both dimensions considering a MS with radial symmetry, as schematized in Figure 1b). To demonstrate the two-axis scanning capability, we present in Figure 3a) the elevation (top) and the azimuthal (bottom) line scanning respectively, to highlight that 150° FoV (supplementary materials S1[47]) is accessible for both scanning axis (see video V1 in supplement materials). These examples of line scanning are realized by fixing the voltage value on the one deflector and scanning the voltage of the second deflector over the entire range at a scanning rate that exceeds the acquisition speed of either our eye or the CCD refreshing frame rate, resulting in an apparent continuous line scans. We prepared a scene (Figure 3b) – bottom) with three different actors located at different angular and depth positions of 1.2m, 2.7m and 4.9m to demonstrate 3D imaging. Due to low laser pulse peak power (about 10mW), we performed our demonstrations in an indoor environment using high reflective suits, considerations of power and losses are addressed in section S2 of supplemental materials[47]. For the demonstration, we choose a visible laser operating at $\lambda = 633nm$, which is very convenient to observe and monitor the deflected beam. After calibrating the system[47], arbitrary -or random access- beam scanning along high-FoV can be realized and arbitrary intensity patterns can be projected by rapidly steering the beam at different locations at very short time intervals, (see video V2 in supplement materials). Figure 3c) shows examples of several scanning profiles implemented to the metasurface beam scanner to project Lissajous curves given by:

$$\theta = A\sin(\alpha t + \Psi)$$

$$\varphi = B\sin(\beta t)$$

$$A = B = 30 \text{ degrees}$$

$$\Psi = 0$$

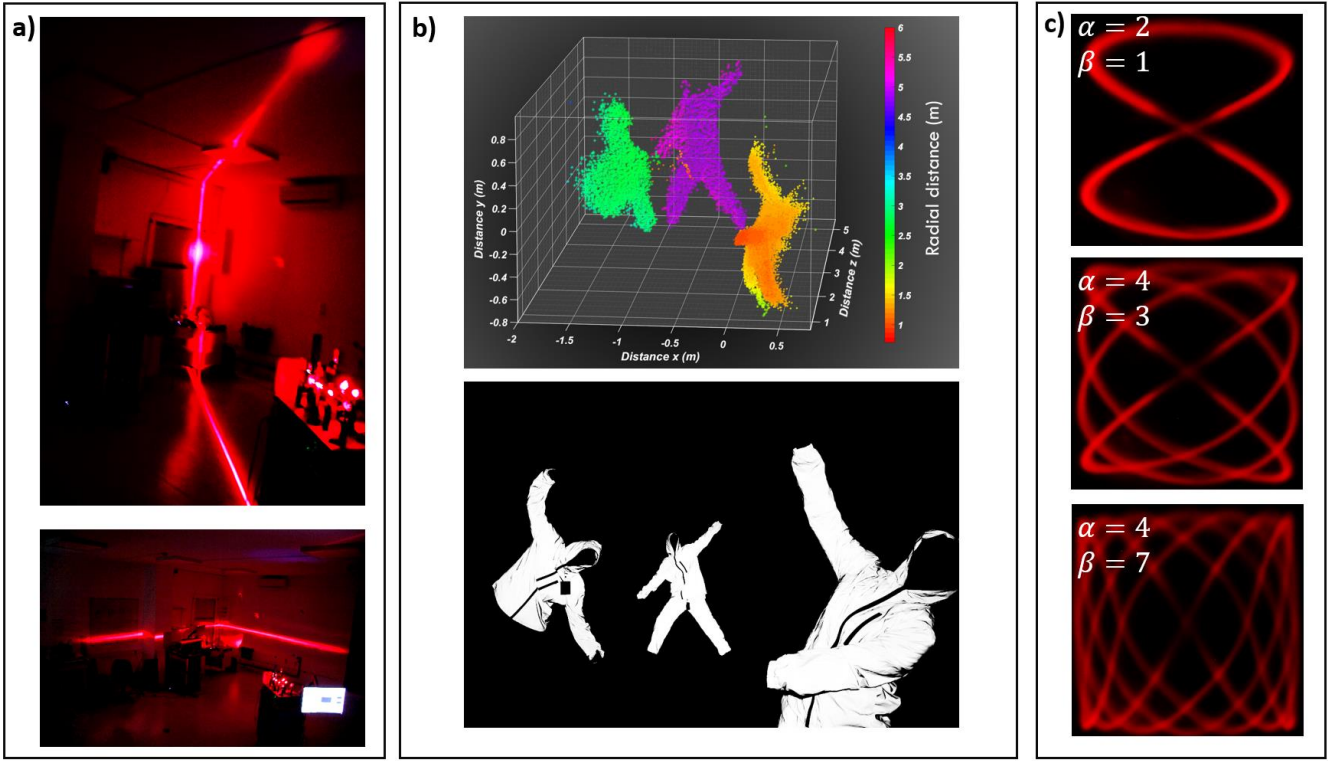

*Figure 3- **a)** LIDAR line scanning of our laboratory room that show the large FoV on both Elevation (top) and Azimuth (bottom) angles. Note the top picture showing a scanning line profile covering the whole range from the ground to the ceiling of the testing room over 150 degrees. **b)** 3D ranging demonstration (top): the scene (bottom) was setup with actors wearing reflective suits positioned in the scene at distance z varying from 1.2 to 4.9m. **c)** Lissajous scanning using deflecting functions as $\theta = A\sin(\alpha t + \Psi)$ and $= B\sin(\beta t)$ for different parameters α and β to illustrate the laser projection capabilities on a fast beam scanning, in a large FoV configuration.*

**Mimicking human peripheral and fovea vision with multi-zone LiDAR imaging**

Previous experiments were performed by focusing the light deflected by the AOD on relatively small metasurfaces (1, 2 and 3mm diameters) using a scanning lens. This configuration favors a small spot (of the order of 50µm) to contain the MS angular divergence to small parametric region, *i.e.* scanning the MS with small spot prevents large overlap with the spatially varying deflecting area. The beam divergence as a function of the metasurface size is provided in the supplementary material S5, indicating that a 3mm device results in a divergence lower than 1.5 deg. Robotic systems interested in reproducing human vision requires peripheral and central vision as illustrated in Figure 4a, where several zones featuring different spatial resolutions are acquired simultaneously. The scene thus needs to be scanned differently according to the zones of interest. To reduce further beam divergence and improve as needed the resolution, it is necessary to increase diameter and complexity of the metasurface and work with fully collimated beams. For this purpose, we realized a cm-size metasurface deflector using nanoimprint lithography (NIL), as shown in Figure 1f)- 1g) (further details on the fabrication are provided in S9). In the latter configuration, the deflector is directly placed after the AOD without utilizing a scanning lens. We specifically designed a large area deflector that achieve moderate 1st order deflection efficiency of ~40% and took advantage of the non-deflected zero order narrow scanning FoV to simultaneously scan two zones with different FoV and resolutions. This demonstration specifically exploits the multi-beam addressing capability of metasurfaces, resulting in a dual mode imaging: i) a high-resolution scanning provided by the near collimated zero order beam deflected by the AOD only, and ii) a large FoV, lower resolution image provided by the 1st order beam deflected by the metasurface. As illustrated in fig 4b-inset, we spatially selected the returned/scattered signal from the different parts of the scene. For this purpose, we used a double-detector monitoring scheme. The first detector collects light from the full numerical aperture (~$2\pi$ solid angle) but it blocks the central small numeric aperture (a beam blocker is placed in front of the detector). The second detector covers only a small NA for the narrow FoV resulting from zero order light scanning (a spatial filter is used to select the observation area). A dual beam metasurface scanning scheme is used to image a scene (fig 4b-top) with two fields of interest: (i) three actors placed at different regions of the space periphery, as measured in Figure 4c) (top) and a highly resolved chessboard-like object placed in the forward direction at a small FoV, measured in Figure 4c) (bottom). The images presented in Figure 4c) correspond to low- and highly resolved imaging, acquired by both detectors simultaneously. Multi-zone scanning with a high resolution forward, and low lateral resolution over a high FoV peripherical vision could be a disruptive solution for addressing the needs of advanced driver assistance systems (ADAS).

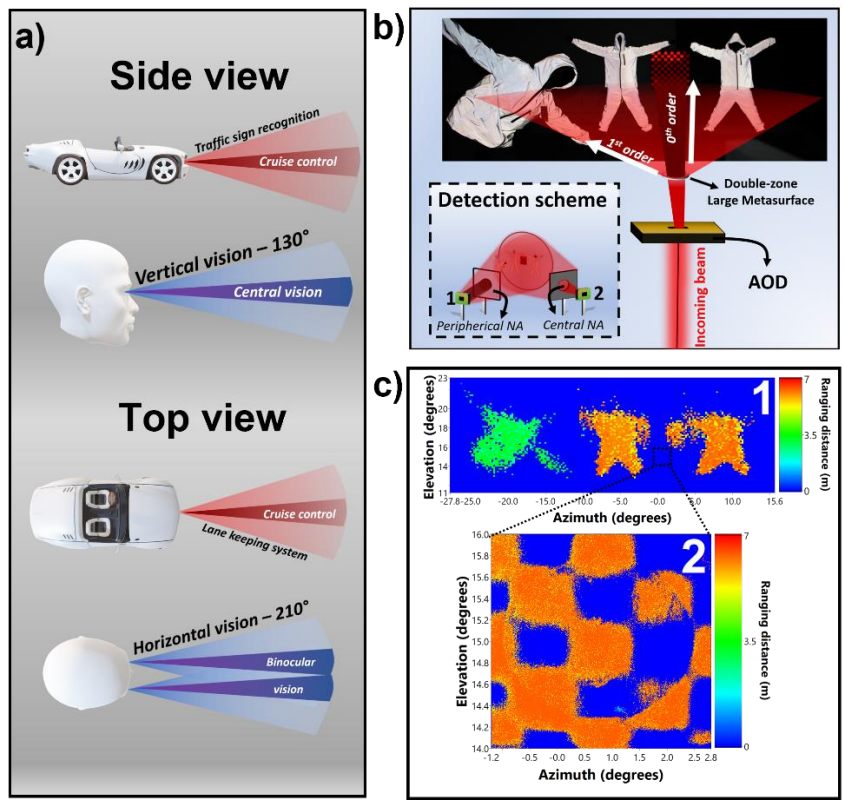

*Figure 4 – **a)** Schematic representation of a human multizone viewing with the concept to be adapted in ADAS systems. Such mimicking characteristics enables double vision for dual-purpose imaging features for high resolution, long range, in the center and lower resolution, bigger FoV, for the peripherical view. **b)** Experimental realization to test the dual zone imaging functionality of the LIDAR system, including dual detection scheme (inset) for simultaneous image multiplexed collection. The central 0th diffraction order beam scans a small area with high resolution directed at the center of the image while the 1st diffracted order scans the whole field. In **c)** (top) we show the result of the scanned scenes described in b). Top represents the LIDAR large FoV ranging image. The image is obtained by blocking the central part of the*

*numerical aperture using an obstacle as sketched in b). The bottom LIDAR ranging high resolution image presents the central part scene captured using the 0th diffraction beam, covering a FoV of about 2deg.*

**High-speed velocimetry and time-series imaging**

To characterize the MHz deflection speed and the possibility of achieving real-time frame rate imaging, we measured the beam deflection speed, *i.e.,* the minimum frequency at which the beam can be re-pointed to a new direction. To do so, we placed highly reflecting tapes on the wall, and measured the amplitude of the backscattered signal for distinct scanning frequencies. We define as "system cutoff frequency" the condition when the amplitude of the reflected signal decays to -3dB point. The measurements were made by considering: i) a single scanner in the azimuth angle (see supplementary materials Figure S5b red curve) and ii) a cascaded system comprised by two orthogonally oriented deflectors, for scanning at both azimuthal and elevation angles, (see supplementary materials Figure S5b) blue curve). The results indicate less than -3dB loss up to around 6MHz and 10MHz for single and double-axis scanning, respectively. We also demonstrate the modulation of a laser beam over an extreme FoV (>140°) at MHz speed and correct imaging with scanning frequency up to 6.25MHz (see supplemental materials Figure S5c)). This corresponds to about two orders of magnitude faster than any other beam pointing technology reported so far. Operating beyond the -3dB loss at higher frequency was also realized, leading to reduced resolution but increased imaging frame rate, up to 1MHz for 1D scanning at 40MHz (see discussion in supplementary materials in section S7).

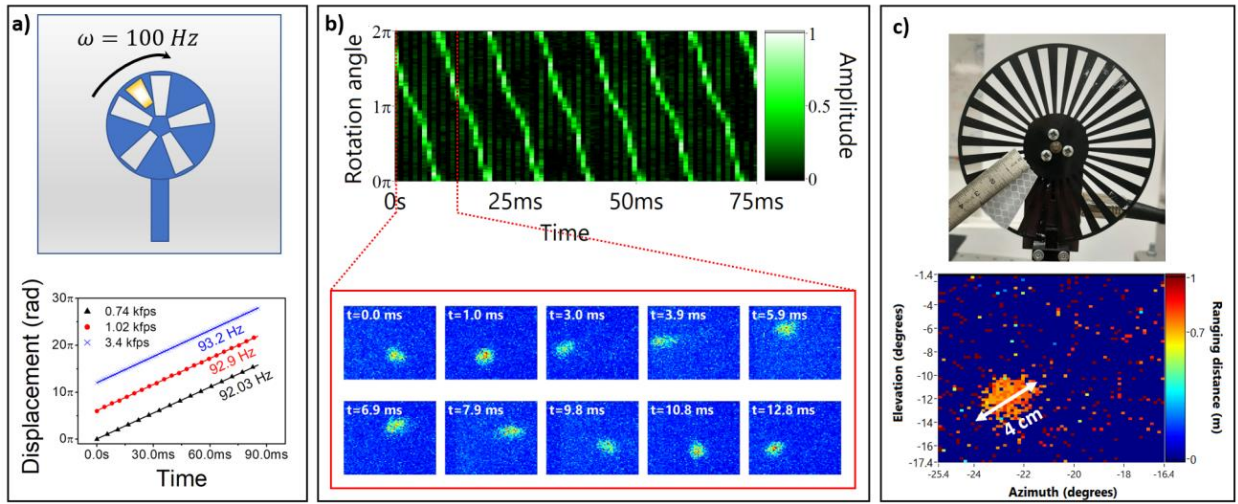

*Figure 5 - Measurement of time-series events. **a)** (Top) Illustration of the scene: a mechanical chopper of was set-up with a nominal speed of 100Hz and some slabs were covered using a reflective tape. (Bottom) Measurement of the rotation speed for three different frame rates. **b)** (top) Normalized intensity map for the radial axis, illustrating the dynamics of the wheel. Note the different slope for the rotation angles around $3\pi/2$ representing a lessening of the speed. (Bottom) Single frame intensity data illustrating various angular positions. **c)** (Top) photography of the chopper and the size of the reflective tape. (Bottom) Ranging image for t=1.0ms and the measurement of the tape from the recovered data.*

Measurements of time events were performed to investigate dynamic imaging. The most convenient dynamic system observable in our laboratory was a spinning chopper composed of a rotating wheel at nominally 100Hz rotation speed. We prepared the scene composed of a chopper, located at 70cm away from the source, decorated with a high reflective tape in one of the mechanical shutters as illustrated in Figure 5a) (top). As described in supplemental table 1, we performed three time-series experiments using acquisition frame rates of 741fps,

1020fps and 3401fps. We tracked the center position of the reflective tape in both the space and time domains by integrating the radial axis of the ranging image from the center of the chopper and fitting a Gaussian curve plotted over the entire $[0,2\pi]$ angular axis (see figure 5a)-bottom). The curves are manually offset by $6\pi$ to differentiate the experiments. All experiments revealed an averaged rotation speed value of 92,71Hz. We attribute the ~7.3Hz difference between the measured and nominal speed of 100Hz to the phase-jitter control mechanism on the chopper. In principle, rotating mechanical shutters are designed with a closed loop circuitry providing an electronic signal that maintains linear rotation speed. Interestingly, displaying time-events on the angular dimension reveals small wobbling wheel imperfection caused by the presence of the reflective tape, resulting in a slowdown at the angles around $3\pi/2$ as evidenced in Figure 5b) (Experiment 2 – 1020fps). One can indeed observe a rotation slope-change during periodic times corresponding to the position of the reflective tape at the bottom (for instance at t=1.0ms/10.8ms in Figure 5b) (bottom-panel)). Using the recovered ranging information, we estimate the size of the tape to 4cm, as illustrated in Figure 5c (bottom). The 1cm difference to the real object (Figure 5c) -top) is due to the high reflectivity of the screws located close to the center and causing additional scattering at the same ranging distance.

These demonstrations open new grounds in high-speed LiDAR imaging, where one can measure very fast objects moving on a large FoV. Employing parameters described on the second row of table 1 (supplementary materials), we achieved a time step of 980 μs, see Figure 5b). An object traveling at the speed of the sound (1234Km/h) at 15m away from the source will take approximately 74ms to cover a 120° FoV. Such supersonic object can be detected within 76 time-series events. Considering the Nyquist limit *i.e.*, 4 time-series to recover the speed, the maximum event-detection can increase up to a speed of 47 Mm/h.

**Experimental methodology**

A collimated beam is sent to an AOD device (AA Opto-electronic DTSXY-400-633) to deflect light at small arbitrary angles, within 49 mrad. The AOD is driven by a voltage-controlled RF generator (AA Opto-electronics DRFA10Y2X-D-34-90.210). The deflected signal is directed to a scanning lens (THORLABS LSM03-VIS) that focuses the light at different transverse positions on the MS. The MS acts as designer-defined passive device to convert the small 2°X2° FoV into an enhanced 150°x150° FoV. ToF is obtained by monitoring the scattered light at each scanned angle using a detector (Hamamatsu C14193-1325SA); and the reconstructed ranging image is built by associating each period ($\frac{1}{f_{rep}}$) to individual pixels and extracting the ToF. A PXI (National Instruments) system is used for data generation, recovery and treatment (more details can be found in section S6 of supplemental materials). The angular scanning of the whole 1D was performed in a single shot, during which we orchestrated pulse repetition, scanning position angles and collection for precise measurement of ToF in the system. With an acquisition scope card of 3Gsamples/s sample rate and considering a rise time on the detector smaller than ~330ps, the maximum z (depth) resolution of single echo per laser shot measurement is about $\Delta z = 5cm$. In Figure 2d) we show the collected raw signal corresponding to the three objects. For ToF recovery we used the derivative of the signal and collected the peak of the differentiated signal. Single pulses were collected (inset Figure 2d)) and separated to evaluate the ToF for each scanned direction and then folded at the scanning frequency to form an image. The fabrication of the different MS have been realized using GaN on sapphire nanofabrication processes. Details are available in the supplementary materials.

**Conclusion**

In conclusion, we realize an ultrafast beam scanning system composed of a fast deflector and a passive metasurface to achieve beam steering at MHz speed over 150x150° FoV, improving the wide-angle scanning rate of mechanical devices by five orders of magnitude. We performed fast steering in one and two angular dimensions and retrieved the associated Time of Flight for ranging measurements. Our approach also offers random-access beam steering capabilities. Multi-zone ranging images mimicking human vision at high frame rate have been realized. The versatility of MS for wavefront engineering could improve the capabilities of simultaneous localization and mapping algorithms. Furthermore, incorporating this system in ADAS could provide an "all-in-one" solution for medium/long range perception, in which the central view scans the front scene, while the peripheral view provides additional sensing for pedestrian safety for example. We finally demonstrated time-event series for imaging at real-time regime (>1kfps and up to MHz frame rate for 1D scanning). Outperforming existing LiDAR technologies, our tool offers perspective for future applications, in particular by participating to reducing the low decision-making latency of robotic and advanced driver-assistance systems.

**Acknowledgements**

This work was financially supported by the European Research Council proof of concept (ERC POC) under the European Union's Horizon 2020 research and innovation program (Project i-LiDAR, grant number 874986), the CNRS prématuration and the UCA Innovation Program (2020 start-up deepTech) and the French defense procurement agency under the ANR ASTRID Maturation program, grant agreement number ANR-18-ASMA-0006-. CK and MS acknowledge inputs of the technical staff at the James Watt Nano-fabrication Centre at Glasgow University.

# Supplementary Information

## Authors

Renato Juliano Martins[1], Emil Marinov[1], M. Aziz Ben Youssef[1], Christina Kyrou[1], Mathilde Joubert[1], Constance Colmagro[1,2], Valentin Gâté[2], Colette Turbil[2], Pierre-Marie Coulon[1], Daniel Turover[2], Samira Khadir[1], Massimo Giudici[3], Charalambos Klitis[4], Marc Sorel[4,5] and Patrice Genevet[1†]

## Affiliations

[1] Université Cote d'Azur, CNRS, CRHEA, Rue Bernard Gregory, Sophia Antipolis 06560 Valbonne, France

[2] NAPA-Technologies, 74160 Archamps, France

[3] Université Côte d'Azur, Centre National de La Recherche Scientifique, Institut de Physique de Nice, F-06560 Valbonne, France

[4] School of Engineering, University of Glasgow, Glasgow, G12 8LT, UK

[5] Institute of Technologies for Communication, Information and Perception (TeCIP), Sant'Anna School of Advanced Studies, Via Moruzzi 1, 56127, Pisa, Italy

† Corresponding Author: Patrice.Genevet@crhea.cnrs.fr

**LiDAR and beam steering videos are available on the Youtube Channel here:**

https://www.youtube.com/channel/UCmezaBH-xOxMjqk3bvqnlAg

## S1. Beam deflection characterization

A solid calibration procedure which correlates the deflection angle with the voltage applied into the AOD RF driver enables reliable beam scanning with low image distortion. The results shown in figure S1a) demonstrate deflection angles varying from -60 to 60 degrees for AOD voltage actuation ranging from 0 to 10V on the 2D deflection system (AA Opto-electronic DTSXY-400-633). The calibration curve in figure S1c) indicates a total FoV of 150 degrees and the green lines denote the absolute maximum deflection angle before the beam shape starts being deformed, accompanied with significant efficiency drop. The displacement of the impact spot on the MS is a function of both the AOD deflection angle and the distance between the AOD and the MS. For calibration purposes and to account for the radial symmetry of the MS, we experimentally constrained the system such that deflections in the range -90 and +90 degrees correspond to +5 and –5 AOD voltage respectively. This 180-degree interval corresponds to the maximum FoV predicted by the sample design. However extreme FoV deflection are not observed in the experiments due to low MS efficiency at extreme angles. If this issue can be addressed by employing a MS designed for high deflection angles. The experimental points were fitted by a third order polynomial function, which provides a continuous and analytical expression allowing global calibration for any arbitrary scanning. Results are shown in Figure 3c) and analytically described hereafter in section S3.

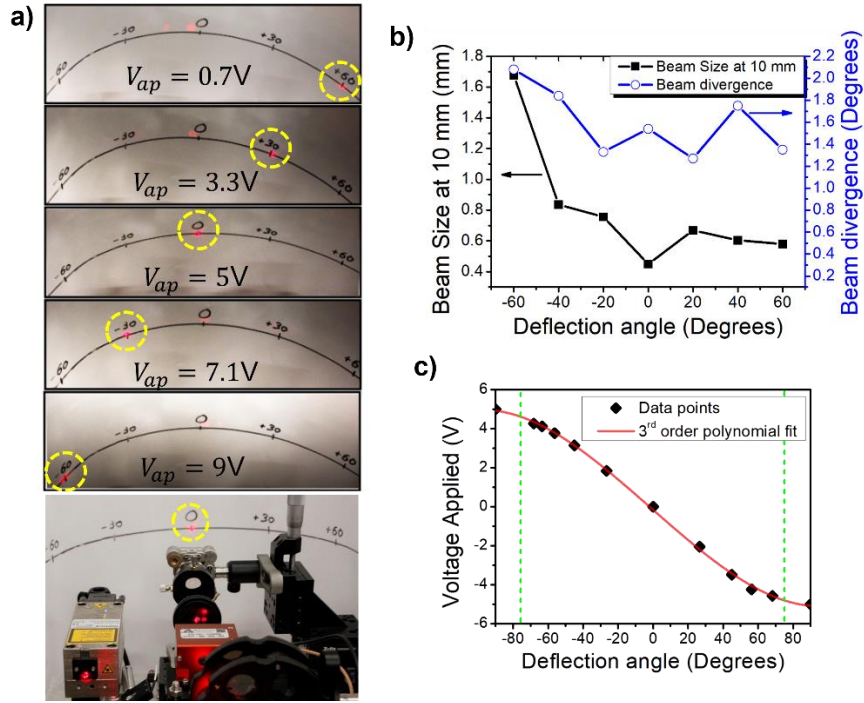

*Figure S1 - Deflection capabilities – In a) we illustrate the beam deflection as function of the applied voltage ranging from 0.7V to 9V, by employing the AOD driver. One can observe some faint small orders of diffraction leaking from the metasurface due to fabrication imperfections, as well as to the overlay of high orders of deflected beam from the metasurface. In the bottom we illustrate the reference of the LIDAR point-of-view. In b) we present the beam size measurement as a function of the deflection angle to characterize the beam divergence. In c) we show the calibration curve as function of the AOD applied voltages. We aligned the system such that the angle of 0 degrees correspond to 0V. This approach allows us to analytically fit the calibration axis of symmetry, R, represented by red curve. Green dashed traces represent the limits in which we were able to measure the deflection; below and above those points important beam distortion occurred and no angle could be accurately measured.*

Moreover, as plotted in fig. S1b), we investigate the beam shape (divergence and size) for several deflection directions, in the range of [-60°, +60°]. These measurements determine the maximum achievable spatial resolution. They also quantitatively show the evolution of the imaging quality for increasing deflection angles. Low divergence is observed when the beam is directed towards the MS center. Working with a scanning lens, i.e. focused beam instead of collimated wavefront, imposes additional distortions. We found a variation in the beam divergence between 1.4 and 2.0 degrees for the beam passing at the center and at -60°, respectively (blue curve, right axis on fig. S1b). The beam waist size increases by around 4 times at -60° due to previously mentioned effects. Note that due to MS design and fabrication imperfections, we observe the presence of a small residual zero$^{th}$ diffraction order located around 0 degree.

### S2. Losses and power distribution

Efficient and low-noise LiDAR sensing requires high power sources to illuminate the scene at a wavelength for which the environment does not add significant absorption/scattering losses. LiDARs utilized for ADAS and/or robotic systems usually operate at IR wavelengths of 905nm/940nm and 1550nm. Prototype system must in principle rely on these wavelengths due to the atmospheric absorption and the lower solar background. Here we are proposing a system operating at conventional visible wavelength for proof of concept demonstration. Visible wavelengths are not optimal for LiDAR but allow an easy visualization of the scanning path. Due to the easy scaling properties of MS, this concept can similarly be exploited at IR frequencies. To characterize the typical optical losses on the modules constituting the system described in Figure 1, we considered all the loss channels, including deflection efficiency of both AOD and MS.

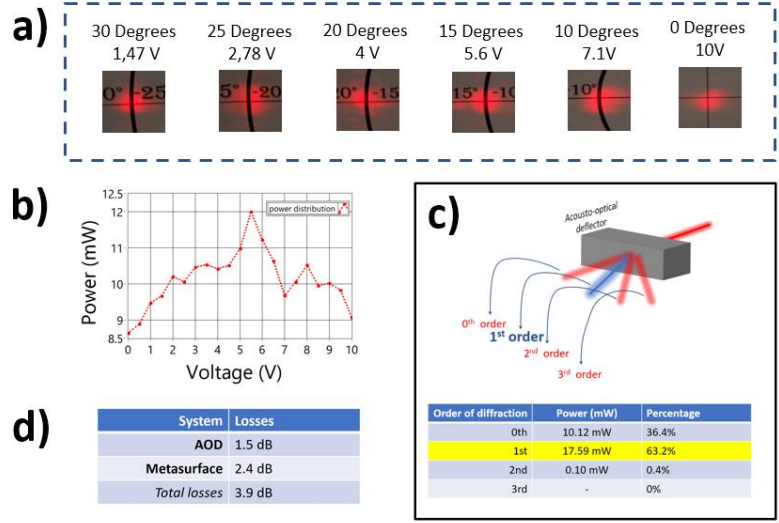

*Figure S 2 – Losses measurements – **a)** Beam profile in function of the deflection angle and the correspondent voltage. When the beam moves towards higher angles, we observe a distortion in the shape evidenced by an elongated shape, at 10V. **b)** Distribution of laser power according to the deflection angle. **c)** Power distribution of the diffracted orders on the AOD when optimizing for the 1st order. **d)** Overall losses measured on the final LiDAR system accounting both lossy AOD and MS.*

In figure S2a) we show the deformation of the beam for the single axis deflection system from Gooch&Housego deflector *AODF 4090-7*. For the 1D case, the highest observed deflection angle was 30° for which we offset the minimal voltage (1.47V). Additionally, in Figure S2b) we show the power distribution along the deflected angles, revealing a variation of 30%. Such variation arises from the decreasing diffraction efficiency versus diffraction angle, i.e. lower performance toward the periphery of the sample. The losses are not only limited by the MS but also by the diffraction efficiency of the AOD. After optimization of the 1st order by adjusting the incident input, we can recover about 63% of the incoming beam intensity on the [1,1] AOD deflected beam, as illustrated in figure S2c). The remaining 37% is leaking principally through the 0th order and secondly to other higher orders. These undesired optical signals are discarded from the measurements using a beam blocker placed between the AOD and the MS. The overall loss of the deflection module is summarized in figure S2d), resulting in a value of around 4dB. Additional loss of 1.5dB should be considered for 2D scanning.

### S3. Calibration of 2-axis scanning

Additional calibration procedures have been realized to address 2D deflection angles, generalizing the curve in Figure S1c). To implement any arbitrary beam scanning, as illustrated in figure 3c), we fabricated a MS with radially symmetric phase profile. Any beam impinging onto the MS will thus behave according to the calibration curve in figure S1c) but by replacing linear coordinate by radial coordinate. The angular position of the beam on the MS plane, recorded by the angle $\theta_{MS}$ as showed in figure S3a) and Figure 1, defines the deflection angles $(\theta_t, \varphi_t)$ in Eq. 3. For accurate calibration, we assume that each point on the MS plane can be translated into a given voltage $V_x$ and $V_y$, applied into the AOD driver as shown in fig. S3a). Converting the voltages $V_x$, $V_y$ into polar coordinates, we define $r$ and $\theta_{MS}$, given by:

$$r = V_x^2 + V_y^2$$
$$\theta_{MS} = \operatorname{atan}\left(\frac{V_y}{V_x}\right) \quad [S1]$$

and:

$$V_x = r\cos(\theta_{MS})$$
$$V_y = r\sin(\theta_{MS}) \quad [S2]$$

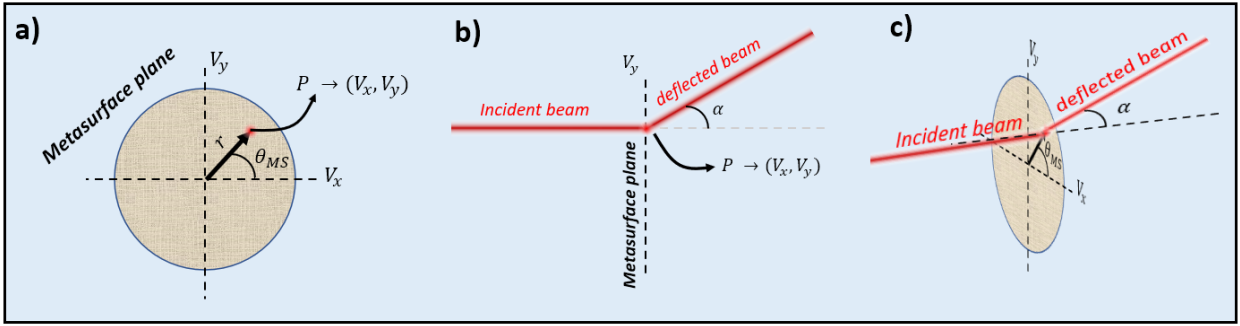

*Figure S1 Illustration of the geometries used to estimate 2D deflection calibration. a) Front view of MS plane to address the angle $\theta_{MS}$ according to the point $P(V_x, V_y)$. b) Side view, where we define a plane of symmetry perpendicular to the MS plane that will deflect the beam to an angle α. c) Perspective view illustrating the connection of the angles.*

Beyond the MS, the beam is deflected to an outgoing general angle $\alpha$ (figure S3 b)), lying within the orthogonal plane to the MS, that acts symmetrically for any $\theta_{MS}$ as illustrated in fig. S3c). Furthermore, the angle $\alpha$ can be easily extracted using the calibration curve in figure S1c), depending on the point $r$. By associating the equivalent spherical coordinates (azimuthal $\theta$ and elevation $\varphi$), we obtain:

$$\theta_{MS} = \operatorname{atan}\left(\frac{\tan\varphi}{\sin\theta}\right)$$
$$\alpha = \operatorname{acos}(\cos\varphi \cos\theta) \quad [S3]$$

Similarly,

$$\theta = \operatorname{atan}(\tan\alpha \cos\theta_{MS})$$
$$\varphi = \operatorname{asin}(\sin\alpha \sin\theta_{MS}) \quad [S4]$$

With these expressions we can transform the scanning coordinates into a system based purely on the voltage control of the AOD driver. After the final conversion from $(\alpha, \theta_{MS}) \to (V_x, V_y)$ from equation S2 and the polynomial fit of the curve on figure S2c), we obtain the 2D calibration curves presented in Figure S4.

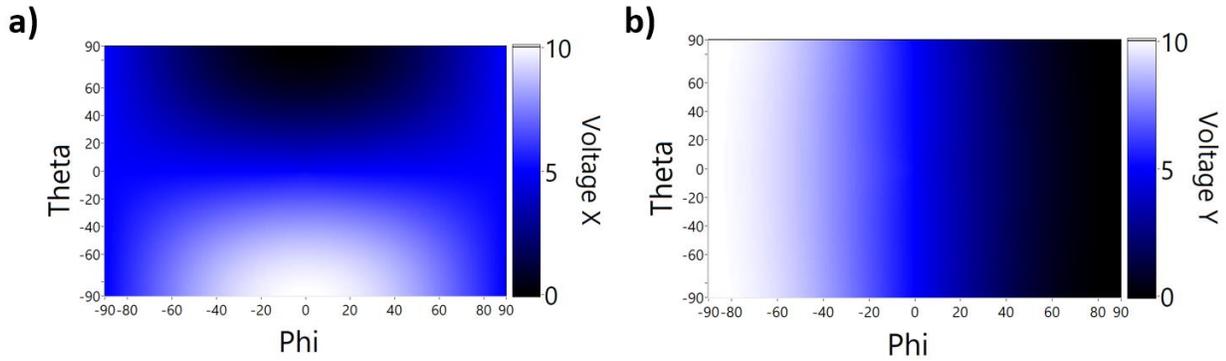

*Figure S2 – Calibration maps for both voltage X in a) and voltage Y in b). These maps are used to associate a given point in the azimuthal θ and elevation ϕ angles to the voltages to be applied into the AOD driver. Such scheme allow us to perform any arbitrary scan as illustrated in figure 3c).*

**S4 Beam modulation speed and imaging quality**

Fast scanning over the limits of AOD brings unwanted beam distortion leading to poor imaging quality, low resolution, and bad beamforming. The beam modulation speed using an AOD is intrinsically limited by the transit time $\tau$ defined by[1]:

$$\tau = \frac{D}{v_\alpha} \qquad [S4]$$

Where $D$ is the beam diammeter (~3mm) and $v_\alpha$ is the acoustic velocity (650m/s). The AOD used in our experiments gives a nominal transit time of $15.4\ ns$ and a nominal scanning frequency of 216 kHz. To further investigate the limit in imaging speed, we firstly estimate the maximum beam deflecting frequency at which we can retrieve a good image quality. To test this scanning speed we performed aback and forth 2-repointing experiment. We defined the threshold frequency as the one for which the repointing signal drops to around 50%. The measurements lead to the observation of frequencies of 6 MHz (blue curve) for the 2D AOD and around 10MHz for 1D AOD beyond the estimated transit time from equation S4. In this regime of fast scanning, small beam deformations are expected and to further investigate the image quality, we performed several intensity imaging tests of a USAF target (figure S5a)) located at 1m away from the source. We observed no image degradation for scanning frequencies lower than 3.33MHz. As shown in the lower row in figure S5c), beyond this frequency, the resolution is drastically reduced indicating that the requested scanning exceeds the transit time inside the AOD and therefore beam deformation is introduced.

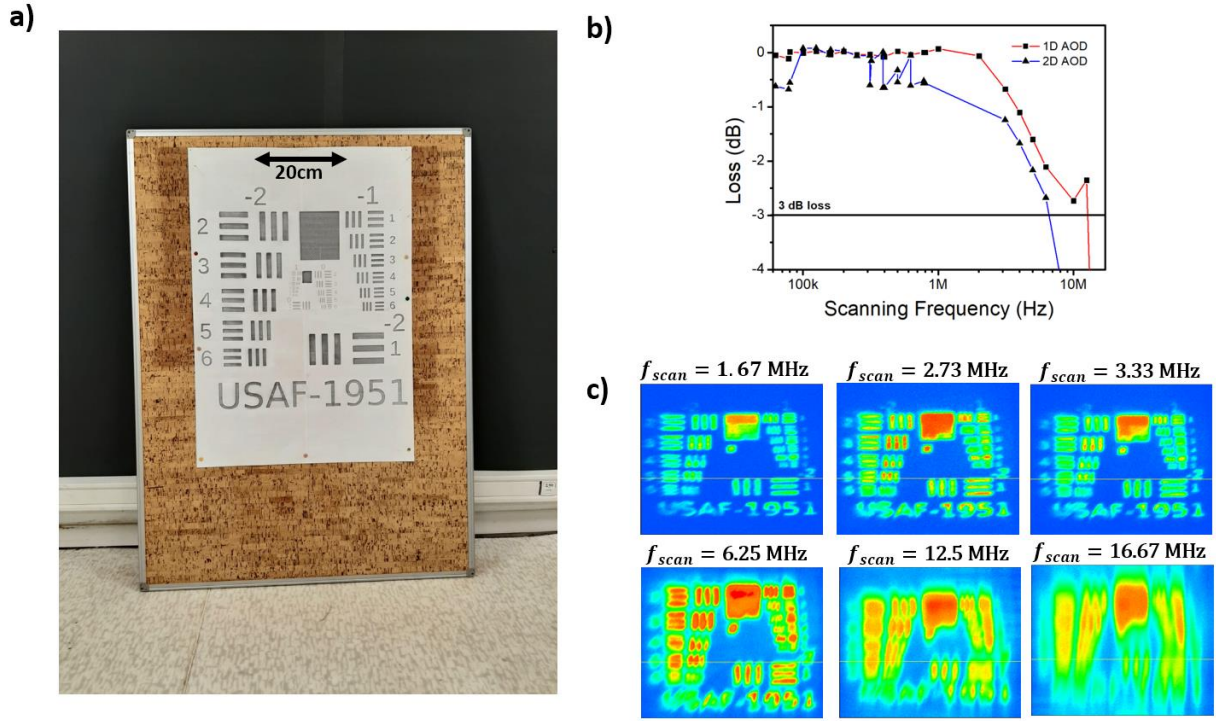

*Figure S3– a) Picture of the target test used to evaluate imaging quality as a function of the scanning frequency. b) Bandwidth response of the scanning system for 1D deflector (blue) and 2D deflector (red). The blackline on the bottom of the graph represents the 3dB threshold point where the amplitude decreases over 50%. c) LiDAR Intensity images showing decreasing image quality. (upper panel) image is still recovered with good resolution (lower panel) degradation of the image as the scanning moves beyond the 3dB threshold.*

**S5 Beam divergence characterization**

The setup illustrated in figure 1a) and 1b), contains a scanning lens which ensures that the beam waist is the same at the focal plane regardless the wave vector. However, the beam impinging on the MS has a physical size, and thus overlays only a certain number of meta-atoms. The section of illuminated meta-atoms is function of the beam size at the focal region. The MS introduces beam divergence that thus depends on the beam size. To analyze the effects of the beam spot on the MS on transmitted beam divergence, we measured the beam diameter (using a CCD camera) as a function of the ranging distance, as illustrated in figure S6a). We collected the beam profile intensity on imaging planes located at various increasing distances from the MS, as show in figure S6b). We extracted the beam profile by defining the central point of the beam and by performing an integration in the angular axis using polar coordinates. The extracted curves and their respective Gaussian fits are illustrated in figure S6c). These data were used to estimate the divergence by performing a linear regression of the beam waist as a function of the distance as illustrated in Figure S6d).

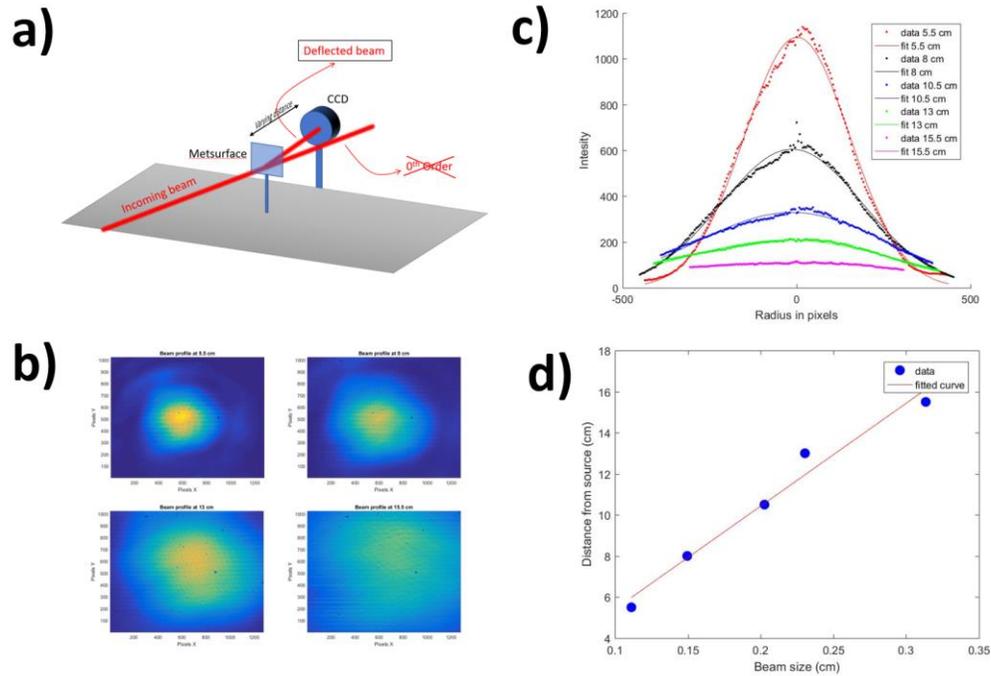

*Figure S6 – Beam divergence setup and measurements for the MS with 1mm of diameter. In **a)** we illustrate the experimental setup employed. We analyze the beam profile for every distance Z from the MS by using a CCD and the results for Z= [5.5,8,13,15.5]cm is illustrated in **b)**. **c)** For each distance we extract the radial profile from the center of the beam and fitted using a Gaussian function. In **d)** we show the linear regression performed given a beam divergence of 4.5°.*

Such experiments and analysis were performed for three different MS diameter sizes (1mm ,2mm and 3mm), while keeping the same MS phase function. Intensity data of the beam profile for different samples are presented at figure S7a) and the compilation is displayed in figure S7b). In fact, and as expected, by increasing the MS size, and thus decreasing the spatial variation of the phase profile for a fixed beam spot, one reduces the divergence of the deflected beam. Considering ADAS and robotic vision applications, a divergence value of 0.25 degrees allows resolving small objects located far away (~200m), (see the patterned dot in figure S7 b)). The graph indicates that such performances are obtained by further increasing the sample diameter up to 3.9mm. Note that such analysis works only when considering the setup including a scanning lens with an input beam of 3mm into the AOD. Any modification of the beam along the setup, including a different choice of optics, would result to different divergence. Furthermore, by changing the optical system, we must reconsider all the calibration curves presented in the figure S4.

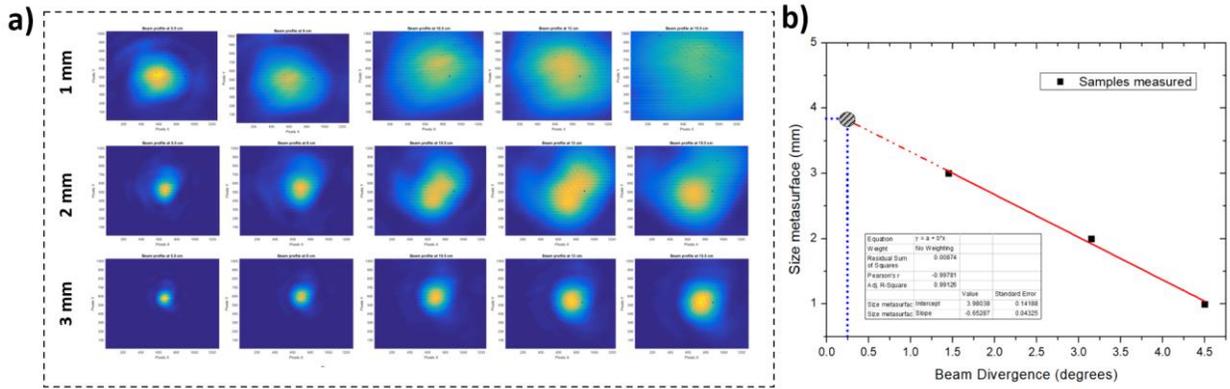

*Figure S7 – a) beam profile to estimate the beam divergence over different ranging z-distances. We performed a Gaussian fit of the radial profile and extracted the FWHM as a function of the distance. Performing a linear regression, we estimated the beam divergence. Such measurements were investigated under different MS diameters and the respective divergence extracted for each sampla, as illustrated in b). Note that a performant LiDAR requires a divergence as low as 0.25 degrees, which would be accessible with a MS of 3.9mm diameter.*

## S6. Data generation, acquisition, and analysis

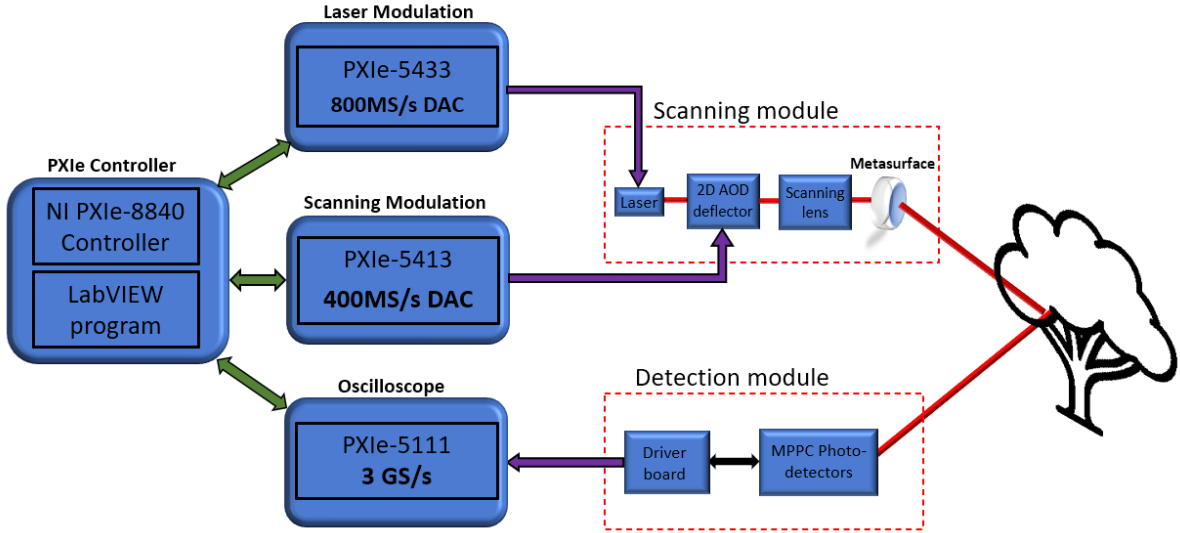

*Figure S8 – Schematics of the logical path to retrieve ranging image. A synchronous PXi system assisted by a graphical user interface in LabView controls different modules to perform various tasks. The scanning modules consists of a laser modulation system and a 2-channels scanning modulation to create single ToF pulses and scanning directions, respectively. The detection module is coupled to an oscilloscope module that acquires, stores and fetches data used for imaging reconstruction.*

To electronically drive the whole system presented in this work, we employed a synchronized PXIe system consisting of different controllers built-in a chassis. The chassis is the foundation of the system, mounted on a rack-type container, providing a common clock and featuring high-speed data-transfer and high speed inter-communication between the modules. This scheme gives PXIe's modules the ability to reach high level of synchronization and low jitter needed for LiDAR systems. The employed modules for the development of the LiDAR presented herein are: (i) a controller - containing the logical part (processor, RAM memory, hard drive), (ii) two frequency generator (FGEN) cards - one to control the scanning, and the other to control the laser, and (iii) an oscilloscope (SCOPE) featuring an analog-to-digital-converter (ADC) to fetch the voltage data from the photodetector. A detailed architecture scheme summarizing the features and the models of all the elements of the LiDAR is provided in figure S8. The module used to drive the laser source is a PXIe-5433 FGEN, generating arbitrary waveforms of +/-12V amplitude, with a maximum sampling rate (SR) of 800MS/s. For controlling the scanning, we employed a PXIe-5413, using a SR of maximum 200MS/s. Finally, for the SCOPE module we used the PXIe-5111, featuring dual channel operation with a shared SR of up to 3GS/S. The use of a double-channel is crucial for enabling the multi-zone imaging capabilities presented in figure 4c). The SCOPE SR defines the depth resolution of the LiDAR: *i.e.*, at 3GS/s, the smallest measurable time is 0.33ns, corresponding to a minimal depth precision of $5\ cm$. To ensure that synchronization remains locked over time and avoid any temporal drift, a SR has been used for each module. Each module's SR has been selected to be an integer multiple of the other modules SR to always ensure an integer number of time synchronous samples. The main constraint is to keep the SR of the SCOPE module as high as possible to assure the maximum ranging precision as discussed in the main paper. In summary, to highly leverage the most performant SRs, we must use 300MS/s for the laser driving FGEN and 150MS/s for the scanner driving FGEN since 3GS/s in the SCOPE is multiple of both these SR.

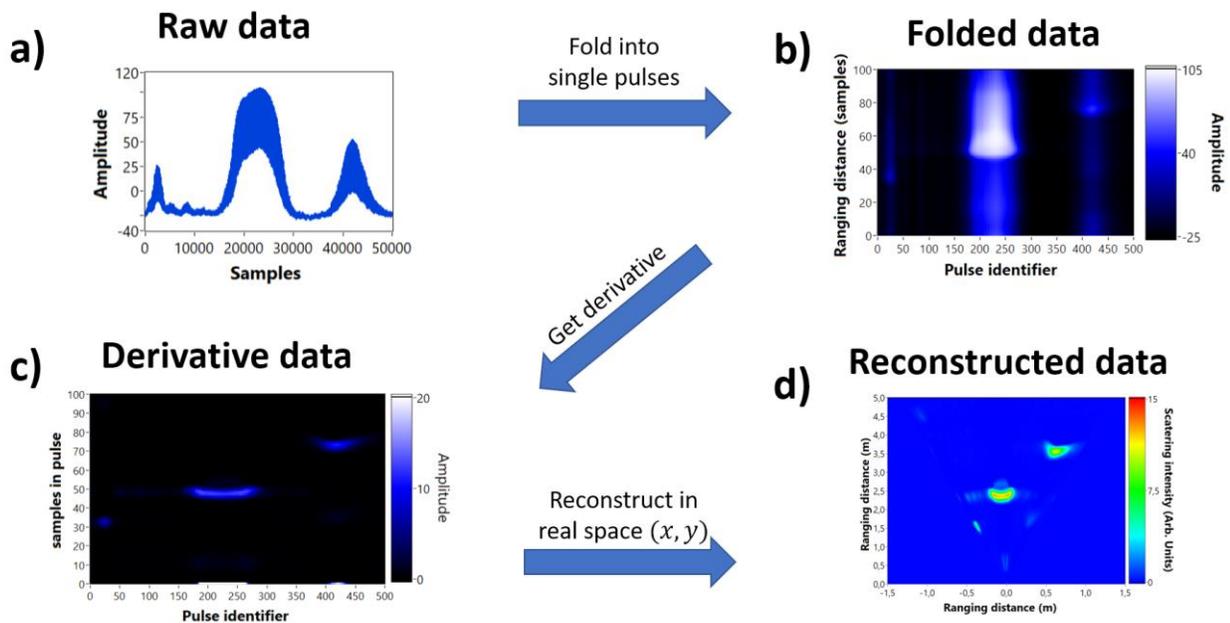

*Figure S9 – Step-by-step of data analysis and imaging retrieving algorithm used in this work. In a) The whole raw data is collected in a single shot measurement corresponding to single ToF pulses from every single line direction of the scan. In b) the data is folded to tag every single direction corresponding to ToF pulses. In c) we perform the derivative of the folded data to find the rising time corresponding to the ToF. Finally in d), we reconstruct the data considering the calibration curve.*

For 2D imaging (1D scanning), data treatment was performed in several steps. After raw signal collection, displayed in Fig S9 a), we folded the data into a matrix (figure S9b)) containing $N \times M$ points, where $N$ is the total number of samples within a single ToF pulse and $M$ indicates the number of resolvable spots in the LiDAR imaging, accounting for the number of pixels. Subsequently, we calculated the derivative of every single $M$ ToF pulses for edge information, as shown in figure S9c). The data is then reconstructed (figure S9d)) by extracting both the intensity and the ToF of every direction, considering the calibrated angles discussed in the last section. Such approach can be extended for 3D imaging by adding another dimension on the scanning.

**S7 – MHz frame rate, 1D scanning**

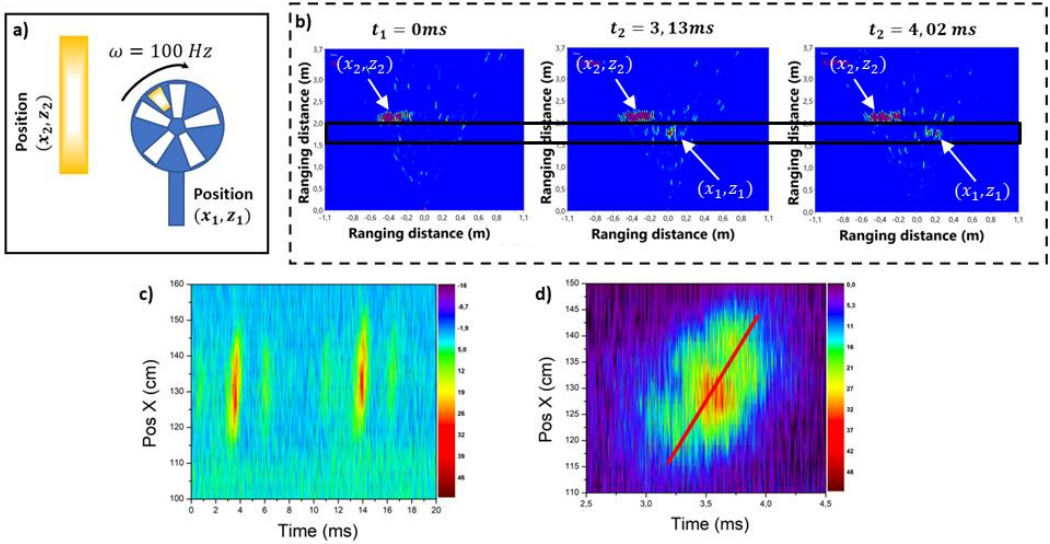

*Figure S10 - Measurement of MHz scanning using the LiDAR. In a) we show the concept of the experiment: an optical chopper with an angular velocity of $100\ Hz$ was place 50cm in front of a high reflective ribbon in the back. A high reflective tape was also placed in one side of the chopper, represented as a golden tape. In b) left we illustrate the ranging image frame cut at the beginning of the experiment at $t = 0\ ms$ which shows the back reflected ribbon at the ranging position of 2m; in the middle, at $t = 3.13\ ms$ we show the moment where the tape passes through the scanning position of the beam at 1.5m represented by a small spot. The plot on the left corresponds to the moment when tape passes close to the end of the scanning line at $t = 4.02ms$. In c) We show a separation of the position 1.5m in the ranging axis which isolates the moving signal only. In graph there are two peaks at $t = 3.5ms$ and $t = 13.5ms$ showing a period of 10ms which corresponds to the 100Hz angular speed. In d) we show a detailed highlight on the first oscillation, the red curve is an illustration of the ribbon speed of around 1400Km/h.*

To demonstrate the ultimate imaging performance of our system, we performed LiDAR imaging of a moving object in two-dimensions (X-Z) by using a single line scanning AOD (GH AODF-4090-7). The fastest available object found in our lab to perform this experiment is a turning optical chopper that was placed on the optical table in the vicinity of a static high reflective ribbon to recover ranging measurement with dynamical and static bodies, as indicated in Figure S10a. The golden tape, at the position $(x_1, z_1)$, represents a moving reflective tape placed on one section of the optical chopper (dynamic object), and the reflective ribbon in the back represents the static object, situated at the position $(x_2, z_2)$. The spinning frequency of the optical chopper was set to the maximum angular speed of $100\ Hz$ allowing fast response characterization. In figure S10b) we present three frame shots taken at different times to illustrate LiDAR's ability to capture moving objects. Particularly, here this dynamic imaging ability is related to the different positions of the reflective tape across the angular scanning range of the LiDAR scene. The image on the left, at $t = 0ms$, corresponds to the instant where the tape does not cross the LIDAR scanning line, in which we observe only the back reflective ribbon located 2m away from source. In the middle panel, at $t = 3.13ms$, the reflective tape passes thought the line of sight of the laser, as indicated by a red amplitude peak around 1.5m away from the source. Finally, at $t = 4ms$, the reflective tape on the chopper moves to the right side of the image, completing the rotation cycle across the LIDAR sight range. Figure S10c) shows the space-time $(x - t)$ LIDAR images obtained after integrated z-pixels around 1.6m, highlighting the movement of the reflective tape. We clearly observe two diagonal peaks delayed by $10\ ms$ and corresponding to the frequency of 100Hz of the oscillation. Furthermore, we can monitor the presence of two small peaks at the side, which correspond to some spurious reflection of the edges of chopper's blades. Isolating the $(x, t)$ zoomed image section at 3.5m, in Figure S10d), we reveal the displacement of the tape along the $x$ direction during a single period of rotation only. For the sake of illustration, we calculated the linear speed of moving tape by placing a diagonal red line on the image, leading to a displacement speed of 30cm/1ms, or ~1400 Km/h. These results demonstrate the ability of the high frame rate imaging LiDAR system to observe fast, even supersonic, moving objects.

## S8. Details on the three dynamic beam scanning experiments

The results presented in the section of high speed and time-series imaging summarized in Figure 5, were repeated for three different configurations according to specifications presented in Table 1. For each measurement we kept the same ranging limit, and we estimated the size of the reflective tape on the chopper. In the first measurement, presented in figure S11a), we setup the imaging to scan at 16.66MHz, aiming to larger number of pixels, 150X150, resulting in a framerate of 741fps. Even with many pixels (scanning points), the estimated size of the tape was slightly bigger than the expected one. This evidence corroborates with the intensity maps showed in figure S5c). Due to the high scanning speed, beyond the 3dB limit, the image quality degenerates. We could still retrieve the dynamics of the rotating chopper but with lower temporal resolution. On the other hand, the 2$^{nd}$ experiment (figure S11 b)), where we kept the scanning speed at 5MHz and we reduced the number of pixels by 70X70, enables retrieving the correct tape size, while keeping a good temporal resolution. Finally, in figure S11c), where the scanning was set to 16.66MHz and the number of pixels was 70X70, we achieved the best temporal resolution of 3kpfs, see central column. The optimal situation is set after compromising between resolution and scanning speed.

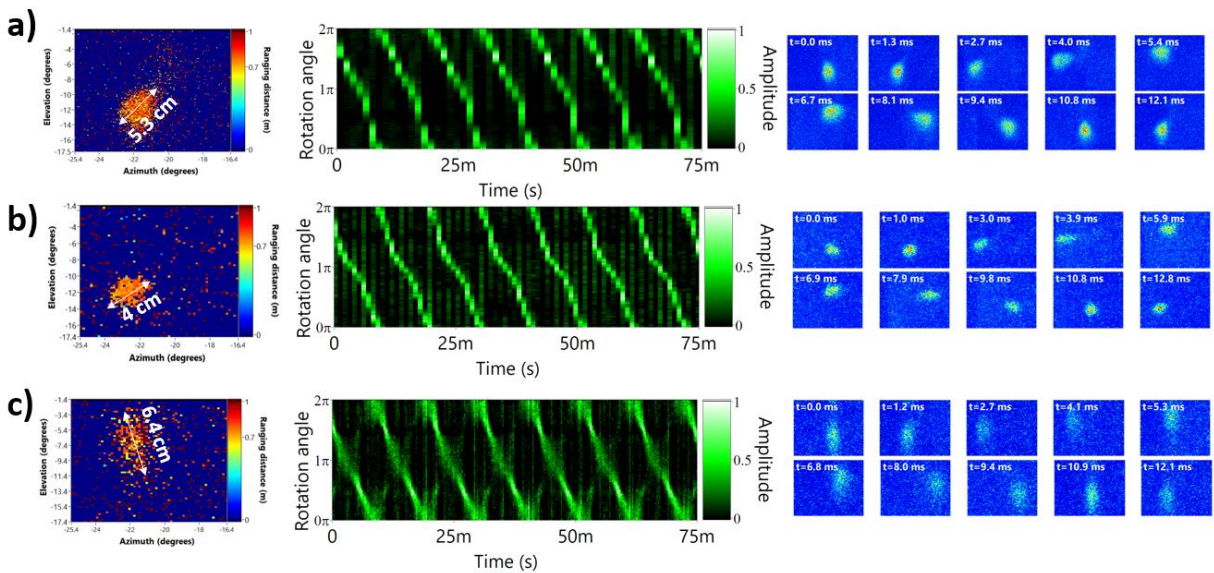

*Figure S41 Dynamic measurements as those presented in Figure 5. Details of experimental conditions for each row are summaried in table 1 for a) 741 fps, b) 1.02kfps and c) 3.4 kfps.*

| Experiment | Framerate | Number of pixels | Time step | speed | Maximum distance | Speed | Total Frames |
|---|---|---|---|---|---|---|---|
| 1 | 741 fps | 150X150 pixels | 1350µs | 16.67 MHz | 8.99m | 92.03Hz | 63 |
| 2 | 1020 fps | 70X70 pixels | 980 µs | 5MHz | 29.97m | 92.9Hz | 87 |
| 3 | 3401 fps | 70X70 pixels | 294 µs | 16.67 MHz | 8.99m | 93.2Hz | 290 |

Table 1 – Parameters of the three experiments to measure the rotation speed. First column is the experiment number. Second column is the number of frames-per-second used in the experiment. Third column summarizes the image resolution. Fourth column represents the time-resolution of each frame collected. Fifth column represent the scanning speed and laser repetition rate used, whose correspondent ranging image is shown in the sixth column. Seventh column is the measured chopper speed, from the curve in fig 7a(bottom). Eight column corresponds to the number of acquired frames using these parameters, such value is limited by the scope card used.

## S9. Metasurface fabrication

The fabrication of the various MS used in these experiments has been realized following two independent processes. For the small components of diameter 1,2 and 3mm we followed our standard GaN metasurface process, consisting of growing a GaN layer on a double side polished (111) Sapphire substrate using Metal-organic chemical vapor deposition (MOCVD) reactor. MOCVD provides accurate thickness control, as well as large-scale uniformity of the GaN layer. We then followed up the fabrication process by employing electron beam lithography, considering a Hydrogen silsesquioxane (HSQ) resist spin-coated onto the GaN. After exposition, the resist pattern is used as an etching mask for the RIE process. Following the GaN RIE etching, the leftover of resist was removed using chemical native oxide removal by dipping the patterned films in a BOE etch. The results of the fabrication are shown in figure S12.

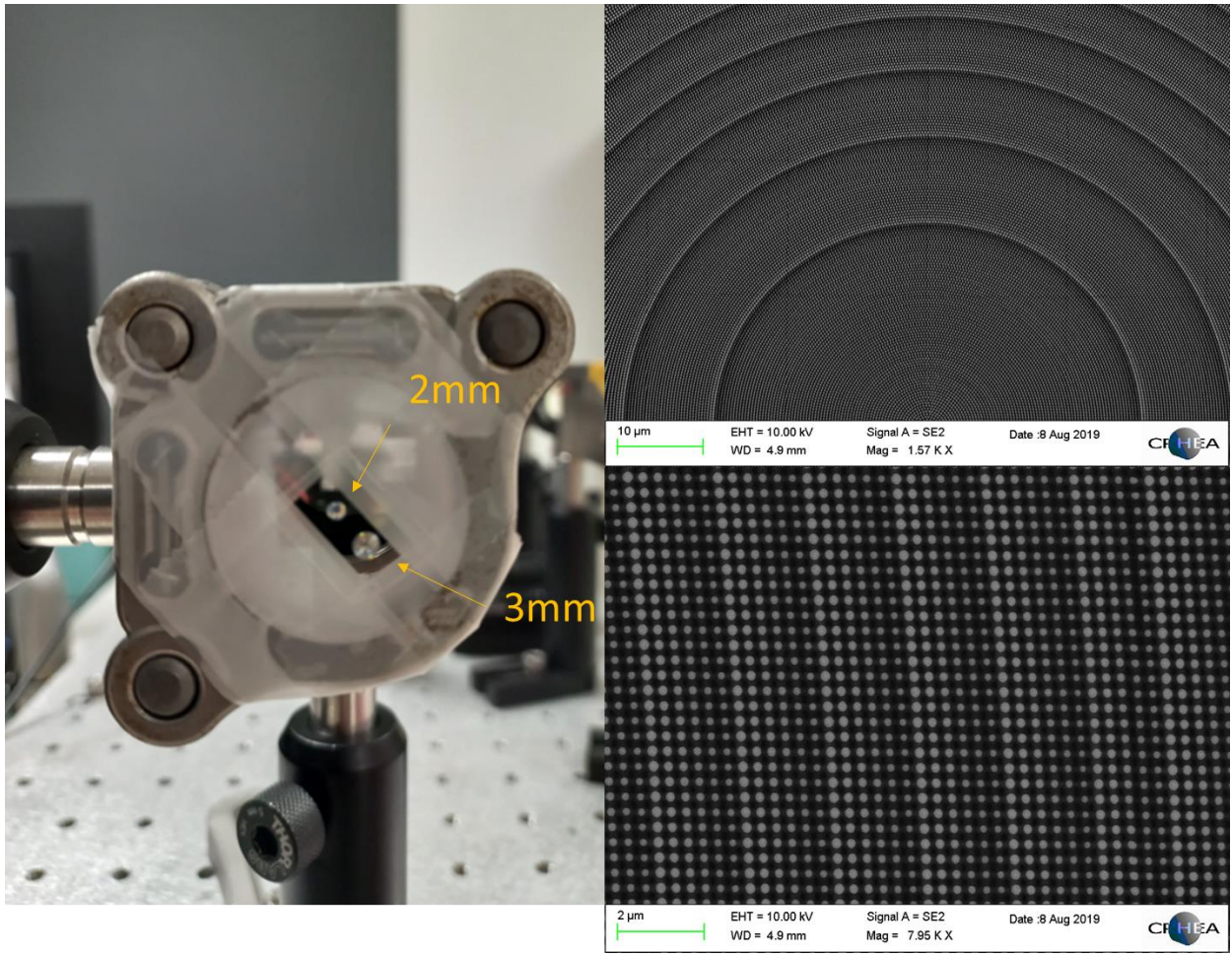

*Figure S52 In a) is a photography of the sample containing two MS deflectors. b) SEM image of the nanopillars showing the size variation to achieve deflection and c) detail on a few nanopillar unit cells to illustrate the phase gradient periodicity;*

The large-area metasurfaces used in dual FoV imaging experiments (see Figure 5) have been fabricated using nano-imprint lithography (NIL). The NIL process starts with the e-beam fabrication of a large area master mold. A soft textured stamp is derived from the master mold, to implement the NIL process. Due to the different fabrication steps, the pattern of the final structure is the negative counterpart of the structure of the master mold. Here, the master mold is designed with holes in order to obtain pillars on GaN etched substrate. The fabrication process is described on figure S13. From the initial master mold, a textured stamp is fabricated, made of the inverse structure (holes in the master mold become pillars in the stamp) (step 1) A sacrificial PMMA layer is coated on a GaN/$Al_2O_3$ substrate. A proprietary $SiO_2$ sol-gel resist is then coated on the PMMA layer and patterned with holes, using the previously fabricated textured stamp (step 2-3). The objective is then to open both resists, to reach the GaN layer of the template. A first etching process is performed using a chlorinated plasma during 5min50s to etch the residual $SiO_2$ sol-gel layer (layer remaining at the bottom of the holes), to reach the PMMA layer. Then, the PMMA is etched during 1min20s. These two etching steps have been implemented by ECR-RIE (step 3-4). Once the pattern is opened, meaning the holes are directly on the GaN layer, a 50 nm think nickel layer is deposited (step 5), on top of the patterns and at the bottom of the holes. The remaining hole structure is then lifted-off: the substrate is immerged in a solvent of the PMMA resist and all resists are removed from the substrate leaving only nanometric nickel disks on the substrate (step 6). Following the lift-off, the GaN template is etched with chlorinated plasma to obtain the desired pillars structure, using the nickel disks as an etching mask (step 7). Finally, the nickel mask is removed by a wet acid etching (step 8).

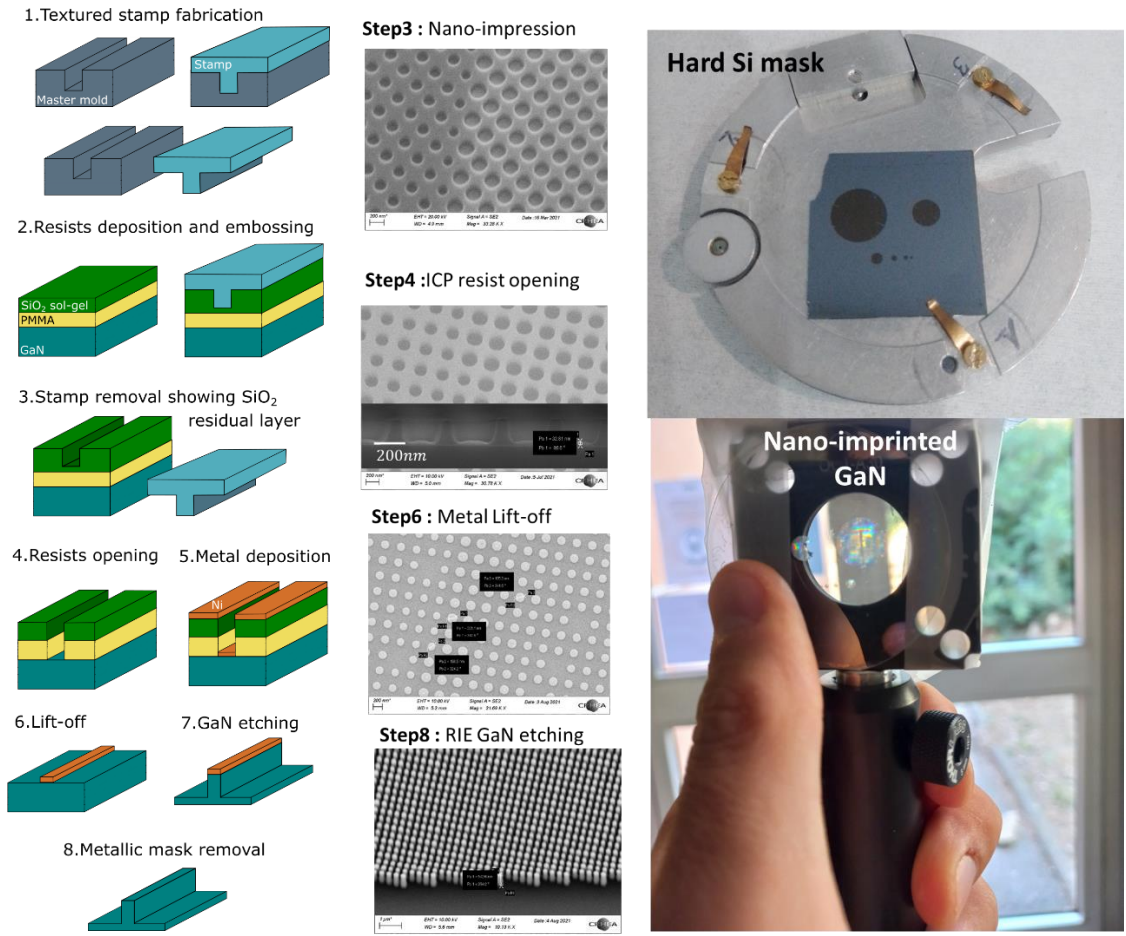

*Figure S63 – Nanoimprinting process to create large area MS used for double zone imaging experiments.*